\documentclass[conference]{IEEEtran}
\usepackage{cite}
\ifCLASSINFOpdf
  \usepackage[pdftex]{graphicx}
  \graphicspath{{figures_small/}{filmstrip_small/}}
  \DeclareGraphicsExtensions{.pdf,.jpeg,.png,.jpg}
\else
\fi
\usepackage[cmex10]{amsmath}
\usepackage{algorithmic}
\usepackage{float}
\usepackage{algorithmic}
\floatstyle{boxed} \floatname{algorithm}{Algorithm}
\newfloat{algorithm}{t}{loa}

\usepackage{array}
\usepackage{mdwmath}
\usepackage{mdwtab}
\usepackage{eqparbox}
\usepackage[font=footnotesize, caption=false]{subfig}
\usepackage{fixltx2e}
\usepackage{url}
\begin{document}
\title{Learning Monocular Reactive UAV Control in Cluttered Natural Environments }

%\author{\IEEEauthorblockN{Michael Shell}
%\IEEEauthorblockA{Robotics Institute\\
%Carnegie Mellon University\\
%Pittsburgh, PA, USA\\
%Email: email@cs.cmu.edu}
%\and
%\IEEEauthorblockN{Andreas Wendel}
%\IEEEauthorblockA{Institute for Computer Graphics and Vision\\
%Graz University of Technology, Austria\\
%Email: wendel@icg.tugraz.at}
%\and
%\IEEEauthorblockN{James Kirk and Montgomery Scott}
%\IEEEauthorblockA{Robotics Institute\\
%Carnegie Mellon University\\
%Pittsburgh, PA, USA\\
%Email: email@cs.cmu.edu}}


\author{\IEEEauthorblockN{St\'ephane Ross\IEEEauthorrefmark{1},
Narek Melik-Barkhudarov\IEEEauthorrefmark{1},
Kumar Shaurya Shankar\IEEEauthorrefmark{1},\\
Andreas Wendel\IEEEauthorrefmark{2},
Debadeepta Dey\IEEEauthorrefmark{1},
J. Andrew Bagnell\IEEEauthorrefmark{1} and
Martial Hebert\IEEEauthorrefmark{1}}
\IEEEauthorblockA{\IEEEauthorrefmark{1}The Robotics Institute\\
Carnegie Mellon University, Pittsburgh, PA, USA\\
Email: \{sross1, nmelikba, kumarsha, debadeep, dbagnell, hebert\}@andrew.cmu.edu}
\IEEEauthorblockA{\IEEEauthorrefmark{2}Institute for Computer Graphics and Vision\\
Graz University of Technology, Austria\\
Email: wendel@icg.tugraz.at}}


\maketitle

\begin{abstract}
%\boldmath
Autonomous navigation for large Unmanned Aerial Vehicles (UAVs) is fairly straight-forward, as expensive sensors and monitoring devices can be employed. In contrast, obstacle avoidance remains a challenging task for Micro Aerial Vehicles (MAVs) which operate at low altitude in cluttered
environments. Unlike large vehicles, MAVs can only carry very light
sensors, such as cameras, making autonomous navigation through obstacles much
more challenging. In this paper, we describe a system that navigates a small
quadrotor helicopter autonomously at low altitude through natural forest
environments. Using only a single cheap camera to perceive the environment, we
are able to maintain a constant velocity of up to 1.5m/s. Given a small set of human pilot demonstrations, we use recent state-of-the-art imitation learning techniques to train a controller that can avoid trees by adapting the MAVs heading. We demonstrate the performance of our system
in a more controlled environment indoors, and in real natural forest
environments outdoors.

\end{abstract}


% \fi
\IEEEpeerreviewmaketitle

%%%%%%%%%%%%%%%%%%%%%%%%%%%%%%%%%%%%%%%%%%%%%%%%%%%%%%%%%%%%%%%%%%%
\section{Introduction}
\label{sec.introduction}

% Paragraph explaining the importance of UAVs in the world for various
% critical tasks
%In the past decade Unmanned Aerial Vehicles (UAVs) have enjoyed considerable
%success in many civil and military applications such as search and rescue,
%monitoring, research, exploration, or mapping. UAVs today come in many
%different shapes and sizes with varying degrees of autonomy.
%% Figure \ref{} shows the large variety of UAVs in current use.
%They range in wingspan from $25$m and payload of several thousand pounds in the case of the %General Atomics Predator and Reaper series
%% \cite{reaper}
% to a wingspan of $7.5$cm and negligible
%payload as in the case of the DARPA Nano Air Vehicles program.
%% \cite{darpanano}
%The
%largest of such UAVs can fly at altitudes of $50000$ feet for $30$ hours while
%the smaller ones can fly at a few meters altitude for a couple of minutes.

In the past decade Unmanned Aerial Vehicles (UAVs) have enjoyed considerable
success in many applications such as search and rescue,
monitoring, research, exploration, or mapping. 
While there has been significant active research in making the
operation of UAVs increasingly autonomous, obstacle avoidance
is still a crucial hurdle. For MAVs
 with very limited payloads it is infeasible to carry
state-of-the-art radars \cite{AmphitechRadar}. Many
impressive advances have recently been made using laser range finders
(lidar)~\cite{bachrach2009autonomous, bryICRA12, scherer2007flying} or
Microsoft Kinect cameras (RGB-D
sensors)~\cite{bachrach2012kinect}. Both sensors are heavy and
active, which leads to increased power consumption and decreased
flight time. In contrast, passive
vision is promising for producing a feasible solution for autonomous MAV
navigation \cite{roberts2012saliency, dey2011saa, wendel2012dense}. 

\begin{figure}[htbp]
	\centering
	\includegraphics[width=0.8\linewidth]{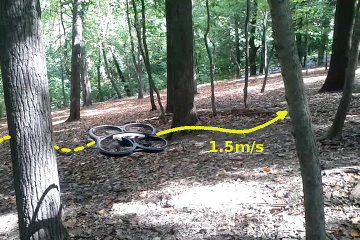}
	\caption{We present a novel method for high-speed, autonomous MAV flight through dense forest areas. The system is based on purely visual input and imitates human reactive control.}
	\label{teaser}
\end{figure}

%Our work is primarily concerned with navigating MAVs which have very low payload capabilities and
%need to fly close to the ground, often in cluttered environments.
Our work is primarily concerned with navigating MAVs that have very low payload capabilities, and
operate close to the ground where they cannot avoid dense obstacle fields.
We present a system that allows the MAV to autonomously fly at high
speeds of up to $1.5$ m/s through a cluttered forest environment (Figure
\ref{teaser}), using passive monocular vision as its only exteroceptive
sensor. We adapt a novel imitation learning technique \cite{ross2011dagger} to
train reactive heading policies based on the knowledge of a human pilot. Visual
features extracted from the corresponding image are mapped to the control
input provided by the expert. In contrast to straightforward supervised
learning~\cite{pomerleau1989alvinn}, our policies are iteratively learned and exploit corrective input at later iterations to boost the overall performance of the predictor, especially
in situations which would not be encountered by a human pilot.
This is an important feature, as the purpose of any reactive controller is to provide a reliable, low-level layer for autonomous control, which works on minimal visual input and can handle situations where 3D mapping~\cite{achtelik2011onboard, wendel2012dense} or high-level trajectory planning~\cite{bellingham2002receding} fails.
Our novel method
is evaluated in a constrained indoor setting using motion capture, as well as
in several forest environments. We successfully avoided more than
$680$ trees during flights over a distance of more than $3$ km.

%%%%%%%%%%%%%%%%%%%%%%%%%%%%%%%%%%%%%%%%%%%%%%%%%%%%%%%%%%%%%%%%%%%
\section{Related Work}
\label{sec.related}

An impressive body of research on control and navigation of MAVs has been published recently. Several state-of-the-art approaches for MAV control would be ideal to fly through a forest, as they feature impressive aggressive maneuvers \cite{mellinger2011minimum} or can even be used for formation flight with large swarms of MAVs \cite{Kushleyev_RSS_2012}. However, these methods still require real-time, accurate
state feedback delivered by a motion-capture system and are therefore unsuitable for our purpose.

% High level related work
% Laser-based:  Nick Roy, Sanjiv Singh, Nicolas Vandapel-Omead Amidi,
The most popular sensors to carry on-board MAVs are laser range finders and
RGB-D sensors, as both deliver quite accurate depth estimates at a high
framerate. Bachrach et al.\ \cite{bachrach2009autonomous} demonstrated using scanning lidars for Simultaneous Localization and Mapping (SLAM) in unknown indoor environments, and Bry et al.\ \cite{bryICRA12}
showed how to use the same sensor for fast flight in indoor environments. The
trend in indoor active sensing has been to use RGB-D sensors~\cite{bachrach2012kinect} that allow more detailed and faster scans. However, in outdoor environments, RGB-D sensors are often not applicable or suffer from very limited range. Therefore, Vandapel et al.\ \cite{Vandapel_2005_4962} proposed outdoor planning approaches in three dimensions for UAV navigation using lidar data, and Scherer et al.\ \cite{scherer2007flying} achieved fast obstacle avoidance using a Yamaha RMax helicopter and a 2-axis scanning lidar. For carrying outdoor lidar systems and the corresponding power supplies, larger and more expensive MAVs than what we aim for are required.

% Vision-based: Randy Beard, Andreas Wendel (for fast reconstruction), Roland
% Siegwart's group, Peter Corke's group, Ashutosh Saxena, Frank Dellaert

% SLAM and reconstruction stuff
Accurate depth estimation and localization is also possible with visual
sensors using stereo cameras\cite{fraundorfer2012stereo} or in a moving
monocular setup~\cite{ccelik2009monocular}. A single, cheap camera is enough
to create sparse~\cite{achtelik2011onboard} or even dense
maps~\cite{wendel2012dense} of the environment. While such
structure-from-motion techniques are already reasonably fast, they are still
too computationally expensive for high-speed flight in a forest. Additionally,
pure forward motion leads to an ill-posed problem when triangulating 3D
points, because the triangulation angle is very small and the resulting position uncertainty is large. 

%Roberts et. al.\ \cite{roberts2012saliency} have focussed on methods
%that produce
%geometric information for planning and navigation in aerial robots and direct
%computation resources to ``salient'' spatial regions which are believed to
%contain crucial information for navigation. 

% Flow
Relatively simple yet efficient algorithms can be derived when imitating animals and insects, who use optical flow for navigation~\cite{srinivasan2011insects}. Beyeler et al.\cite{beyeler2009vision} as well as Conroy et al.\
\cite{conroy2009opticalflow} implemented systems which exploit this fact and
lead to good obstacle avoidance results. Later, Lee et al.\ \cite{lee2010two}
proposed to use a probabilistic method
of computing optical flow for more robust distance calculation to obstacles for
MAV navigation. Optical flow based controllers navigate by balancing flow on
either side. However flow captures
richer scene information than these controllers are able to use. We
embed flow in a data-driven framework to automatically derive a controller
which exploits this information.

% More vision-based stuff
Most closely related to our approach are approaches which learn motion policies and depth from input data. Michels et al.\ \cite{michels2005high} demonstrated driving a remote-controlled
toy car outdoors using monocular vision and reinforcement learning. Hadsell et al.~\cite{hadsell2009learning} showed in the LAGR project how to recognize and avoid obstacles within complex outdoor environments using vision and deep hierarchical networks. Bill et al.\ \cite{bills2011autonomous} used the often orthogonal
structure of indoor scenes to estimate vanishing points and navigate a MAV in corridors by going towards the dominant vanishing point. We extend those approaches and employ a novel imitation learning technique that allows us to find a collision-free path through a forest despite the diverse appearance of visual input.

%%%%%%%%%%%%%%%%%%%%%%%%%%%%%%%%%%%%%%%%%%%%%%%%%%%%%%%%%%%%%%%%%%%
%\input{system.tex}

%%%%%%%%%%%%%%%%%%%%%%%%%%%%%%%%%%%%%%%%%%%%%%%%%%%%%%%%%%%%%%%%%%%
%\section{Learning the Reactive Controller through Imitation}
\section{Learning to Imitate Reactive Human Control}

Visual features extracted from camera input provide a rich set of information
that we can use to control the MAV and avoid obstacles. We leverage
state-of-the-art imitation learning
techniques~\cite{pomerleau1989alvinn,schaal1999imitation,ratliff2006ioc,argall2009survey,ross2011dagger}
to learn such a controller. These data-driven approaches allow us to directly
learn a control strategy that mimics an expert pilot's choice of actions based
on demonstrations of the desired behavior, i.e., sample flight trajectories
through cluttered environments.

\subsection{Background}
The traditional imitation learning approach is formulated as a standard
supervised learning problem similar to, e.g., spam filtering, in which a
corpus of training examples is provided. Each example consists of an
environment (an image acquired by the MAV) and the action taken by an expert
in that same environment. The learning algorithm returns the policy that
best mimics the expert's actions on these examples. The classic successful
demonstration of this approach in robotics is that of \emph{ALVINN}
(Autonomous Land Vehicle in a Neural Network) \cite{pomerleau1989alvinn} which
demonstrated the ability to learn highway driving strategies by mapping camera
images to steering angles.

%%% ***talk about IOC techniques below?***

While various learning techniques have been applied to imitation
learning \cite{schaal1999imitation,ratliff2006ioc,argall2009survey}, these
applications all violate the main assumption made by statistical learning approaches that the learner's predictions (actions) do not affect the distribution of inputs/states. As shown in previous work
\cite{ross2010smile} and confirmed in the MAV setting here, ignoring the
effects of the learner on the underlying state distribution leads to serious
practical difficulties and poor performance. For example, during typical pilot
demonstrations of the task, trees are avoided fairly early and most training
examples consist of straight trajectories with trees on the side. However, since the
learned controller does not behave perfectly, the MAV encounters situations
where it is directly heading for a tree and is closer than the human ever
was. As the hard turns it needs to perform in these cases are nonexistent in
the training data, it simply cannot learn the proper recovery behavior.

Theoretically, \cite{ross2010smile} showed that even if a good
policy that mimics the expert's actions well on the training examples is learned, when
controlling the drone, its
divergence from the correct controls could be much larger (by as much as a
factor $T$, when executing for $T$ timesteps) due to the change in
environments encountered under its own controls.

Fortunately, Ross et al. \cite{ross2011dagger} proposed a simple iterative
training procedure called DAgger (DAtaset Aggregation), that addresses this issue and provides improved performance guarantees. Due to its
simplicity, practicality and improved guarantees, we use this approach
to learn the controller for our drone. While \cite{ross2011dagger}
demonstrated successful application of this technique in simulated
environments (video game applications), our experiments show that this
technique can also be successfully applied on real robotic platforms. We
briefly review the DAgger algorithm below.

\subsection{The DAgger Algorithm}

DAgger trains a policy that mimics the expert's behavior through multiple iterations
of training. Initially, the expert demonstrates the task and a first policy $\pi_1$
is learned on this data (by solving a classification or regression
problem). Then, at iteration $n$, the learner's current policy $\pi_{n-1}$ is
executed to collect more data about the expert's behavior. In our
particular scenario, the drone executes its own controls based on $\pi_{n-1}$, and
as the drone is flying, the pilot provides the correct actions to perform in
the environments the drone visits, via a joystick. This allows the
learner to collect data in new situations which the current policy might visit, but which
were not previously observed under the expert demonstrations, and learn the
proper recovery behavior when these are encountered. The next policy
$\pi_{n}$ is obtained by training a policy on all the training data
collected over all iterations (from iteration 1 to $n$). This is iterated for
some number of iterations $N$ and the best policy found at mimicing the expert
under its induced distribution of environments is returned. See \cite{ross2011dagger} for details.
%This algorithm is summarized in Algorithm~\ref{algDAgger}.

%\begin{algorithm}
%\begin{algorithmic}
%\STATE Initialize $D \leftarrow \emptyset$, $\pi_1$ to query the expert and execute the expert's action.
%\FOR{$n=1$ \textbf{to} $N$}
%\STATE Sample new trajectories by executing $\pi_{n}$.
%\STATE Get dataset $D_n$ of the visited information states associated with the corresponding expert's actions.
%\STATE Aggregate dataset: $D = D \cup D_n$.
%\STATE Train $\pi_{n+1}$ to minimize loss on $D_n$ 
%\ENDFOR
%\STATE \textbf{Return} best $\pi_n$ at mimicking expert under its induced trajectories.
%\end{algorithmic}
%\caption{DAgger for imitation learning~\cite{ross2011dagger}. \label{algDAgger}}
%\end{algorithm}

The intuition is that, over the iterations, we collect a set of inputs the learner is likely to observe during its execution based on previous experience (training iterations), and obtain the proper behavior from the pilot in these situations. \cite{ross2011dagger} showed theoretically that after a sufficient
number of iterations, DAgger is guaranteed to find a policy that when executed at test time, mimics the expert at least as well as how it could do on the aggregate dataset of all training examples. Hence the divergence in controls is not
increased by a factor $T$ as in the traditional supervised learning approach
when the learned policy controls the MAV.

For our application, we aim to learn a linear controller of
the drone's left-right velocity that mimics the pilot's behavior to avoid
trees as the drone moves forward at fixed velocity and altitude. That is,
given a vector of visual features $x$ from the current image, we compute a
left-right velocity $\hat{y} = w^\top x$ that is sent to the drone, where $w$
are the parameters of the linear controller that we learn from the training
examples. To optimize $w$, we solve a ridge regression problem at each
iteration of DAgger. Given the matrix of observed visual features $X$ (each
row is an observed feature vector), and the vector $y$ of associated
left-right velocity commands by the pilot, over all iterations of training, we
solve $w = (X^\top X + R)^{-1} X^\top y$, where $R$ is a diagonal matrix of
per-feature regularization terms. We choose to have individual
regularization for different types of features,
which might represent different fractions of the feature vector $X$, so that every type
contributes equally to the controls. In other words, we regularize each feature of a
certain type proportionally to the number of features of that type. Features
are also normalized to have mean zero and variance 1, based on all the
observed data, before computing $w$, and $w$ is applied to normalized features
when controlling the drone.

\subsection{Using DAgger in Practice}

\begin{figure}%[htbp]
	\centering
	\includegraphics[width=0.65\linewidth]{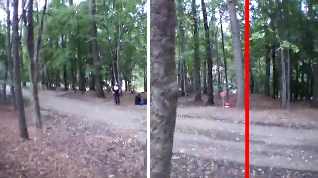}
	\caption{One frame from MAV camera stream. The white line indicates the
          current yaw commanded by the current DAgger policy $\pi_{n-1}$ while the
          red line indicates
          the expert commanded yaw. In this frame DAgger is wrongly heading
          for the tree in the middle while the expert is providing the correct
          yaw command to go to the right instead. These expert controls are
          recorded for training later iterations but not executed in the
          current run.}
	\label{current_hci}
\end{figure}

Figure \ref{current_hci} shows the DAgger control interface used to provide correct
actions to the drone. Note that, at iteration $n$, the
learner's current policy $\pi_{n-1}$ is in control of the MAV and the expert just
provides the correct controls for the scenes that the MAV visits. The expert
controls are recorded but not executed on the MAV. This results in some
human-computer-interaction challenges: 

1) After the first iteration, the pilot must be able to
provide the correct actions without feedback of how the drone would react to
the current command. While deciding whether the drone should go left or right is easy, it can be hard to input the correct magnitude of the turn the drone should perform without feedback. In particular, we
observed that this often makes the pilot turn excessively when
providing the training examples after the first iteration. Performance can degrade
quickly if the learner starts to mimic these imperfect actions. To address this issue, we provided partial feedback to the pilot by
showing a vertical line in the camera image seen by the pilot that would slide
left or right based on the current joystick command performed. As this line
indicated roughly where the drone would move under the current command, this
led to improved actions provided by the pilot (Figure
\ref{current_hci}).

2) In addition to the lack of feedback, providing the correct actions in
real-time after the first iteration when the drone is in control can be hard
for the pilot as he must react to what the drone is doing and not what he
expects to happen: e.g., if the drone suddenly starts turning towards a
tree nearby, the pilot must quickly start turning the other way to indicate
the proper behavior. The pilot's reaction time to the
drone's behavior can lead to extra delay in the correct actions specified by
the pilot. By trying to react quickly, he may provide imperfect actions as well. This becomes more and more of an issue the faster the drone is flying. To address this issue, we allowed the pilot to indicate
the correct actions offline while the camera stream from the drone is replayed
at slower speed (3 times slower than real-time), using the interface seen in Figure
\ref{current_hci}. By
replaying the stream slower, the pilot can provide more accurate commands and react more quickly to the drone's behavior. The faster we're flying, the slower the trajectory could be replayed to provide good commands in time. 

3) The third challenge is that DAgger needs to collect data for all
situations encountered by the current policy in later iterations. This would
include situations where the drone crashes into obstacles if the current
policy is not good enough. For safety reasons, we allow the pilot to take over or
force an
emergency landing to avoid crashes as much as possible. This
implies that the training data used is not exactly what DAgger
would need, but instead a subset of training examples encountered by
the current policy when it is within a ``safe'' region. Despite this modification, the
guarantees of DAgger still hold as long as a policy that can stay within this
``safe'' region can be learned.

\subsection{Features}

Our approach learns a controller that maps RGB images from the on-board camera to control commands. This requires mapping camera images to a set of features which can be used by DAgger.
% This is a standard approach when using machine learning methods in computer vision and a vast variety of features has been used and documented in literature.
These visual features need to provide indirect information about the three-dimensional structure of the environment. Accordingly, we focused on features which have been shown to correlate well with depth cues such as those in~\cite{michels2005high}, specifically Radon transform statistics, structure tensor statistics, Laws' masks and optical flow. 

We compute features over square windows in the image, with a 50\% overlap between neighboring windows. The feature vectors of all windows are then concatenated into a single feature vector. The choice of the number of windows is driven primarily by computational constraints. A $15\times 7$ discretization (in $x$ and $y$ respectively) performs well and can be computed in real-time. 

\paragraph{Radon features (30 dim.)} The Radon
transform~\cite{helgason1999radon} of an image is computed by summing up the
pixel values along a discretized set of lines in the image, resulting in a 2D
matrix where the axes are the two parameters of a line in 2D, $\theta$ and
$s$.  We discretize this matrix in to $15\times 15$ bins, and for each angle $\theta$ the two highest values are recorded. This encodes the orientations of strong edges in the image.
\paragraph{Structure tensor statistics (15 dim.)} At every point in a window
the structure tensor~\cite{harris1988corner} is computed and the angle between
the two eigenvectors is used to index in to a 15-bin histogram for the entire window. The corresponding eigenvalues are accumulated in the bins. In contrast to the Radon transform, the structure tensor is a more local descriptor of texture. Together with Radon features the texture gradients are captured, which are strong monocular depth cues \cite{wu2004perceiving}.
\paragraph{Laws' masks (8 dim.)} Laws' masks~\cite{davies1997} encode texture intensities. We use six masks obtained by pairwise combinations of one dimensional masks: (L)evel, (E)dge and (S)pot. The image is converted to the YCrCb colorspace and the LL mask is applied to all three channels. The remaining five masks are applied to the Y channel only. The results are computed for each window and the mean absolute value of each mask response is recorded.
\paragraph{Optical flow (5 dim.)} Finally, we compute dense optical flow~\cite{Werlberger2010} and extract the minimum and maximum of the flow magnitude, mean flow and standard deviation in $x$ and $y$. Since optical flow computations can be erroneous, we record the entropy of the flow as a quality measure. Optical flow is also an important cue for depth estimation as closer objects result in higher flow magnitude.
%Since optical flow computations can sometimes be meaningless, e.g., on large uniform regions, a quality flag is also recorded, resulting in seven features.

Useful features must have two key properties. First, they need to be computed fast enough.
Our set of features can be computed at 15~Hz using the graphics processing unit (GPU) for dense optical flow computation. Although optical flow is helpful, we show in our experiments that removing this feature on platforms without a GPU does not harm the approach significantly.
Secondly, the features need to be sufficently invariant to changes between training and testing conditions so that the system does not overfit to training conditions. We therefore refrained from adding color features, as considerable variations under different illumination conditions and confusions between trees and ground, as well as between leaves and grass, might occur. An experimental evaluation of the importance of every feature is given in the next section, along with a detailed evaluation.

In addition to visual features, we append 9 additional features: low pass filtered history of previous commands (with 7 different exponentially decaying time periods), the sideways drift measured by the onboard IMU, and the deviation in yaw from the initial direction. Previous commands encode past motion which helps to smooth the controller. The drift feature provides context to the pilot's commands and accounts for motion caused by inertia. The difference in yaw is meant to reduce drift from the initial orientation.

%%%%%%%%%%%%%%%%%%%%%%%%%%%%%%%%%%%%%%%%%%%%%%%%%%%%%%%%%%%%%%%%%%%
\section{Experiments}
\label{sec.experiments}

%\subsection{Setup}

We use a cheap, commercially available quad-rotor helicopter, namely the Parrot ARDrone, as our airborne platform. The ARDrone weights only $420$g and has a size of $0.3 \times 0.3$m. It features a front-facing camera of $320 \times 240$ pixels and a $93\deg$ field of view (FOV), an
ultrasound altimeter, a low resolution down-facing camera and an onboard
IMU. The drone's onboard controller stabilizes the drone and allows
control of the UAV through high-level desired velocity commands
(forward-backward, left-right and up-down velocities, as well as yaw rotation) and can reach a maximum velocity of about $5$m/s. Communication is based on WiFi, with camera images streamed at about $10-15$Hz. This allows us to control the
drone on
a separate computer that receives and processes the images from the drone, and
then sends commands to the drone at around $10$Hz.

%% Note: for the final we could add a link to our own ROS driver here...

% \begin{figure}[!htbp]
%   \centering
%   \mbox{
%     \subfloat[Motion-capture arena indoor setup]{\label{indoor_setup}\includegraphics[width=1.0\linewidth]{figures/drone_indoor_setup.png}}
%   }
%   \mbox{
%     \subfloat[Outdoor groundstation setup]{\label{outdoor_setup}\includegraphics[width=1.0\linewidth]{figures/outdoor_setup.png}}
%   } 
%   \mbox{
%   \subfloat[Long range wifi antenna]{\label{robot_on_map}\includegraphics[width=1.0\linewidth]{figures/wifi_outdoors.png}}
%   }
% \caption{Indoor and outdoor setup}
% \label{setup_figure}
% \end{figure}

\subsection{Indoor Experiments}

\begin{figure}%[htbp]
	\centering
	\subfloat{ \includegraphics[width=0.41\linewidth]{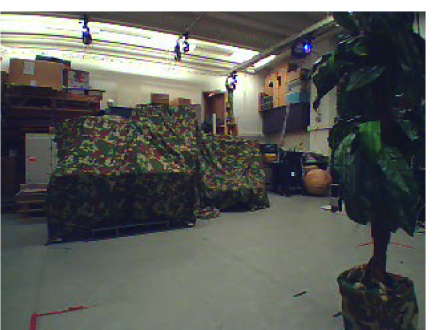} }
	\subfloat{ \includegraphics[width=0.49\linewidth]{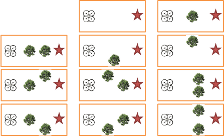} }
        \caption{Left: Indoor setup in motion capture arena with fake plastic trees and camouflage in background. Right: The 11 obstacle arrangements used to train Dagger for every iteration in the motion capture arena. The star indicates the goal location.}
	\label{dagger_indoor_setup}
\end{figure}

%\begin{figure}[htbp]
%	\centering
%	\includegraphics[width=0.7\textwidth]{figures/37MocapToGoal.png}
%	\caption{The drone controller learns to move straight towards the goal location while swerving to avoid trees in the way.}
%	\label{dagger_mocap_to_goal}
%\end{figure}

\begin{figure}%[htbp]
	\centering
	\includegraphics[width=0.28\linewidth]{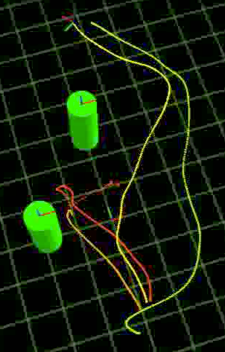}
	\includegraphics[width=0.7\linewidth]{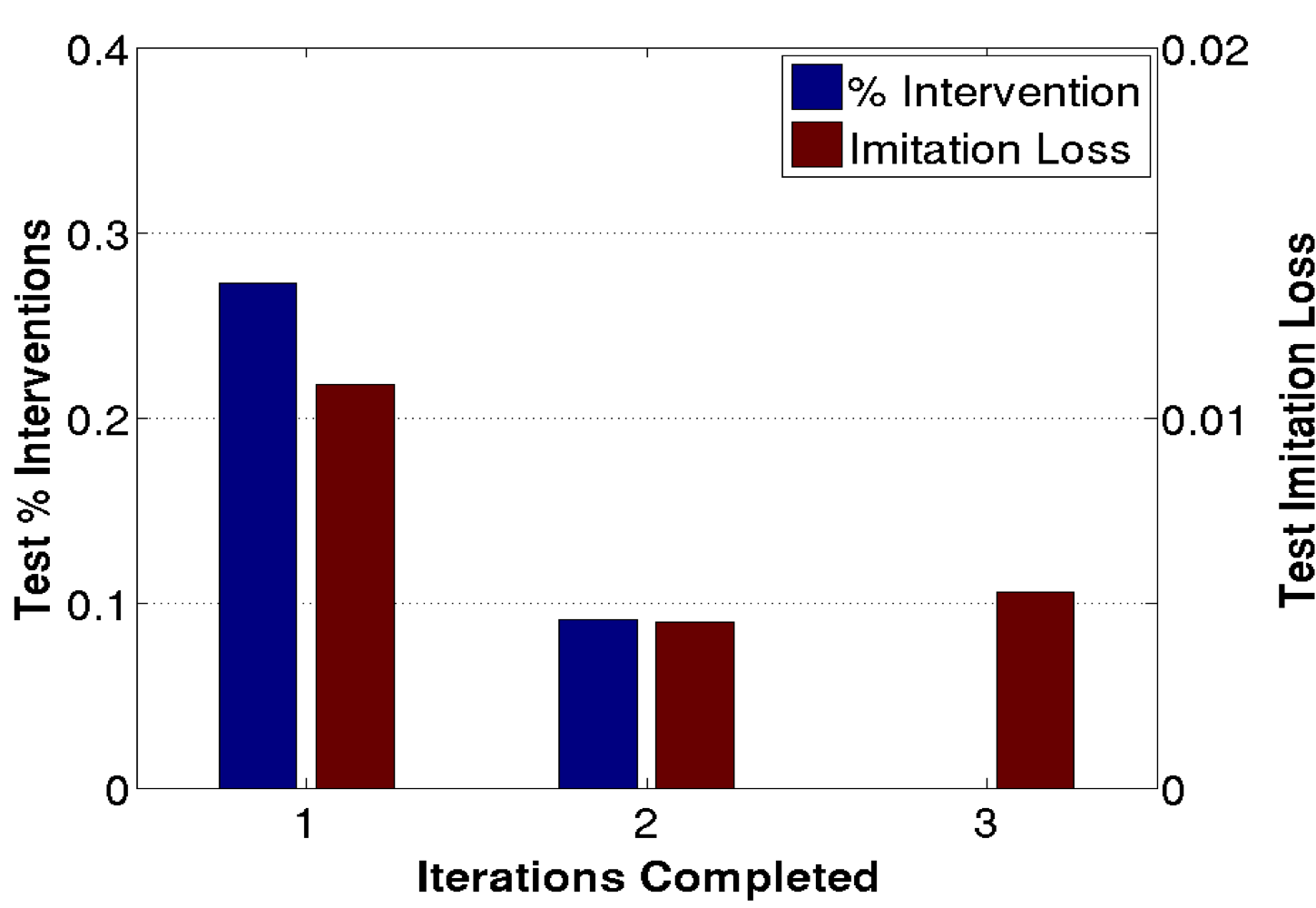}
        \caption{Left: Improvement of trajectory by DAgger over the iterations. The rightmost green trajectory is the pilot demonstration. The short trajectories in red \& orange show the controller learnt in the 1$^{st}$ and 2$^{nd}$ iterations which fail. The 3$^{rd}$ iteration controller succesfully avoids both obstacles and is similar to the demonstrated trajectory. Right: Percentage of scenarios the pilot had to intervene and the imitation loss (average squared error in controls of controller to human expert on hold-out data) after each iteration of Dagger. After 3 iterations, there was no need for the pilot to intervene and the UAV could successfully avoid all obstacles}
	%\caption{The rightmost trajectory is the initial demonstrated trajectory by the human expert flying the UAV looking at the front-facing camera stream. The short trajectories in red show the results of the controllers learnt in the first and second iterations which collide. The third iteration controller no longer collides and is similar to the demonstrated trajectory.}
	\label{dagger_traj_comparison}
\end{figure}

We first tested our approach indoors in a motion capture arena. We use fake indoor trees as obstacles and camouflage to hide background clutter (Figure \ref{dagger_indoor_setup}). While this is a very controlled environment that lacks many of the complexities of real outdoor scenes, it allows us to obtain better quantitive results to determine the effectiveness of our approach.

%\begin{figure}[htbp]
%	\centering
%	\includegraphics[width=0.95\linewidth]{figures/dagger_interventions.png}
%	\caption{Percentage of scenarios the pilot had to intervene and the imitation loss (average squared error in controls of controller to human expert on hold-out data) after each iteration of Dagger. After 3 iterations, there was no need for the pilot to intervene and the UAV could successfully avoid all obstacles.}
%	\label{dagger_interventions}
%\end{figure}
The motion capture system is only used to track the drone and adjust its heading so that it is always heading straight towards a given goal location. The drone moves at a fixed altitude and forward velocity of 0.35m/s and we learn a controller that controls the left-right velocity using DAgger over 3 training iterations. At each iteration, we used $11$ fixed scenarios to collect training data, including $1$ scenario with no obstacles, $3$ with one obstacle and $7$ with two obstacles (Figure \ref{dagger_indoor_setup}).

Figure \ref{dagger_traj_comparison} qualitatively compares the trajectories taken by the MAV in the mocap arena after each iteration of training on one of the particular scenarios. In the first iteration, the green trajectory to the farthest right is the demonstrated trajectory by the human expert pilot. The short red and orange trajectories are the trajectories taken by the MAV after the $1^{st}$ and $2^{nd}$ iterations were completed. Note that both fail to avoid the obstacle. After the $3^{rd}$ iteration, however, the controller learned a trajectory which avoids both obstacles. The percentage of scenarios where the pilot had to intervene for the learned controller after each iteration can be found in Figure \ref{dagger_traj_comparison}. The number of required interventions  decreases between iterations and after 3 iterations, there was no need to intervene as the MAV successfully avoided all obstacles in all scenarios.

\subsection{Feature Evaluation}

%In the current implementation we transmit images from the drone to the
%workstation at 15-30~Hz, and compute the entire set of features at 15~Hz. For our current flight speed of approx. 1~m/s and
%command rate of 10~Hz, this is fast enough. We verified the robustness of the
%features by training and testing under very different conditions as
%illustrated in Figure~\ref{fig:generalize}.

%\begin{figure}%[htbp]
%	\centering
%\begin{tabular}{cc}
%	\includegraphics[width=0.7\linewidth]{figures/14train.png} \\
%	\includegraphics[width=0.7\linewidth]{figures/14test.png} 
%\end{tabular}
%	\caption{Example of different observations used at training time (top) and test time (bottom) to validate the robustness and generalization power of the visual features.}
%	\label{fig:generalize}
%\end{figure}

\begin{figure}%[htbp]
	\centering
	\includegraphics[width=0.65\linewidth]{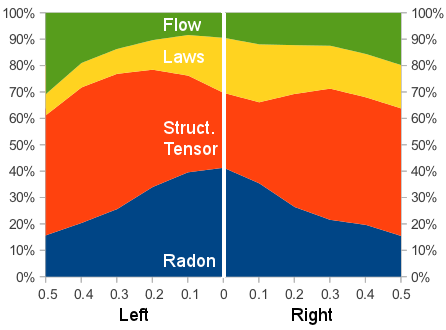}
	\caption{Breakdown of the contribution of the different features for different control prediction strengths, averaged over 9389 datapoints. Laws and Radon are more significant in cases when small controls are performed (e.g. empty scenes), whereas the structure tensor and optical flow are responsible for strong controls (e.g. in cases where the scene contains an imminent obstacle). A slight bias to the left can be seen, which is consistent to observations in the field. Best viewed in color.}
	\label{fig-weights}
\end{figure}

\begin{figure*}
	\centering
	\subfloat[Radon]{ \includegraphics[width=0.18\textwidth]{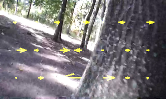} } 
	\subfloat[Structure Tensor]{ \includegraphics[width=0.18\textwidth]{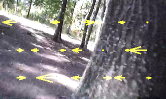} }
	\subfloat[Laws]{ \includegraphics[width=0.18\textwidth]{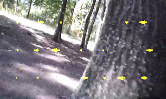} } 
	\subfloat[Flow]{ \includegraphics[width=0.18\textwidth]{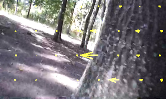} }
	\subfloat[Combined Features]{ \includegraphics[width=0.18\textwidth]{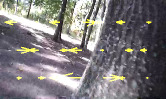} }

	\caption{Visualization of the contribution of the different features to the predicted control. The overall control was a hard left command. The arrows show the contribution of a given feature at every window. Structure tensor features have the largest contribution in this example, while Radon has the least.}
	\label{fig:feat-arrows}
\end{figure*}

After verifying the general functionality of our approach, we evaluate the benefit of all four feature types. 
An ablative analysis on the data shows that the structure tensor features are most important, followed by Laws features. Figure \ref{fig-weights} shows how the contribution of different features varies for different control signal strengths. 
%This visualization shows that the feature types have varying significance for different predictions (which typically correspond to different types of scenes).
Optical flow, for example, carries little information in scenes where small commands are predicted. This is intuitive since in these cases there are typically no close obstacles and subsequently no significant variation in optical flow. In fact, removing the optical flow feature on platforms without sufficient computational capabilities only results in a 6.5\% increase in imitation loss.  

Anecdotally, Figure~\ref{fig:feat-arrows} shows the contribution of each of the features at different window centers in the image. While structure tensor features mainly fire due to texture in the background (indicating free space), strong optical flow vectors correspond to very close objects. In this example the predictor commands a hard left turn (numerical value: 0.47L on a scale of [0,1]), and  all visual features contribute to this. Consistent with the above analysis, the contribution of the structure tensor was greatest (0.38L), Laws masks and optical flow contribute the same (0.05L) while Radon features provide the least contribution (0.01L). In this particular example, the non-visual features actually predict a small right command (0.02R).

%\begin{figure*}[htbp]
%	\centering
%\begin{tabular}{cc}
%	\includegraphics[width=0.2\textwidth]{frame17-radon} 
%	\includegraphics[width=0.2\textwidth]{frame17-harris} 
%	\includegraphics[width=0.2\textwidth]{frame17-laws} 
%	\includegraphics[width=0.2\textwidth]{frame17-flow} 
%	\includegraphics[width=0.2\textwidth]{frame17-all}
%\end{tabular}

\subsection{Outdoor Experiments}

%\begin{figure}[!t]
%\centering
%\includegraphics[width=2.5in]{dagger_improvement}
%\caption{The above plots show that as Dagger progresses the number of
%  interventions per unit distance and imitation loss with respect to the human
%  expert goes down.}
%\label{fig.dagger_improvement}
%\end{figure}

%\begin{figure}[!t]
%\centering
%\includegraphics[width=2.5in]{overhead_map}
%\caption{Shows the actual trajectories taken by Dagger at each iteration for
%  different starting locations. The trajectories are formatted by iteration
%  number. As iterations increase they can avoid obstacles for a longer distance. }
%\label{fig.overhead_map}
%\end{figure}

%\begin{figure}[!t]
%\centering
%\includegraphics[width=2.5in]{system_diagram}
%\caption{Shows our overall system architecture and the various latencies.}
%\label{fig.system_diagram}
%\end{figure}

After validating our approach indoors in the motion capture arena, we
conducted outdoor experiments to test in real-world scenarios. As we could not use the motion capture system outdoors to make the drone head towards a specific goal location, we made the drone move forward at a fixed speed and aimed for learning a controller that would swerve left or right to avoid any trees on the way, while maintaining the initial heading. Training and testing were conducted in forest areas while restraining the aircraft using a light-weight tether. 

We performed two experiments with DAgger to evaluate its performance in different regions, one in a park with relatively low tree density, and another in a dense forest.

\subsubsection{Low-density test region}

%\begin{figure}[htbp]
%	\centering
%	\includegraphics[width=0.7\linewidth]{figures/48testSite.png}
%	\caption{The two test sites and the main training site for the sparser dataset}
%	\label{dagger_sites}
%\end{figure}

The first area is a park area with a low tree density of approximately 1 tree per $12 \times 12$m, consisting mostly of large trees and a few thinner trees. In this area we flew at a fixed velocity of around $1$m/s, and learned a heading (left-right) controller for avoiding trees using DAgger over 3 training iterations. This represented a total of $1$km of flight training data. Then, we exhaustively tested the final controller over an additional $800$m of flight in the training area and a separate test area. 

\begin{figure}%[!htbp]
	\centering
	\includegraphics[width=0.29\linewidth]{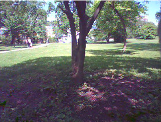}
        \includegraphics[width=0.29\linewidth]{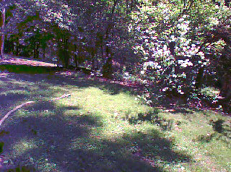}	
        \includegraphics[width=0.33\linewidth]{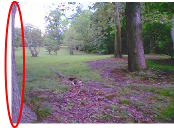}
        \caption{Common failures over iterations. While the controller has problems with tree trunks during the 1$^{st}$ iteration (left), this improves considerably towards the 3$^{rd}$ iteration, where mainly foliage causes problems (middle). Over all iterations, the most common failures are due to the narrow FOV of the camera where some trees barely appear to one side of the camera or are just hidden outside the view (right). When the UAV turns to avoid a visible tree in a bit farther away it collides with the tree to the side.}
	\label{dagger_failures}
\end{figure}

Qualitatively, we observed that the behavior of the drone improved over iterations. After the first iteration of training, the drone sometimes failed to avoid large trees even when they were in the middle of the image in plain view (Figure~\ref{dagger_failures}, left). At later iterations however, this rarely occured. On the other hand, we observed that the MAV had more difficulty detecting branches or bushes. The fact that few such obstacles were seen in the training data, coupled with the inability of the human pilot to distinguish them from the background, contributed to the difficulty of dealing with these obstacles. We expect that better visual features or improved camera resolution might help, as small branches often cannot be seen in $320 \times 240$ pixel images.

\begin{figure}%[!htbp]
        \centering
%	\subfloat[After iter. 1]{ \includegraphics[width=0.24\linewidth,trim=210 250 190 300,clip]{plots/piechart-typecrash-iter1-may5} }~ \hfill
%	\subfloat[After iter. 2]{ \includegraphics[width=0.24\linewidth,trim=210 250 190 300,clip]{plots/piechart-typecrash-iter2-may5} }~ \hfill
%	\subfloat[After iter. 3]{ \includegraphics[width=0.24\linewidth,trim=210 250 190 300,clip]{plots/piechart-typecrash-iter3-may5} } \\
    \subfloat[Pilot]{ \includegraphics[width=0.215\linewidth,trim=210 250 190 350,clip]{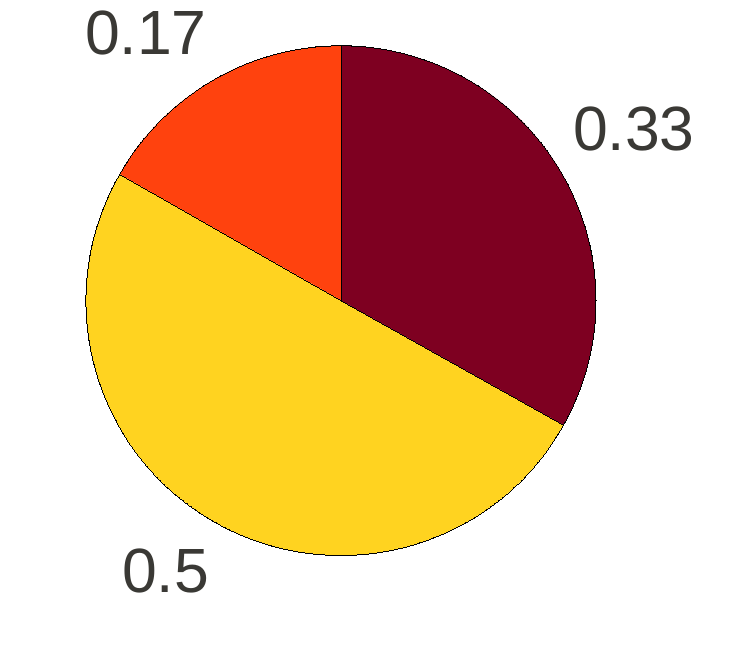} }~ 
	\subfloat[After iter. 1]{ \includegraphics[width=0.215\linewidth,trim=210 250 190 350,clip]{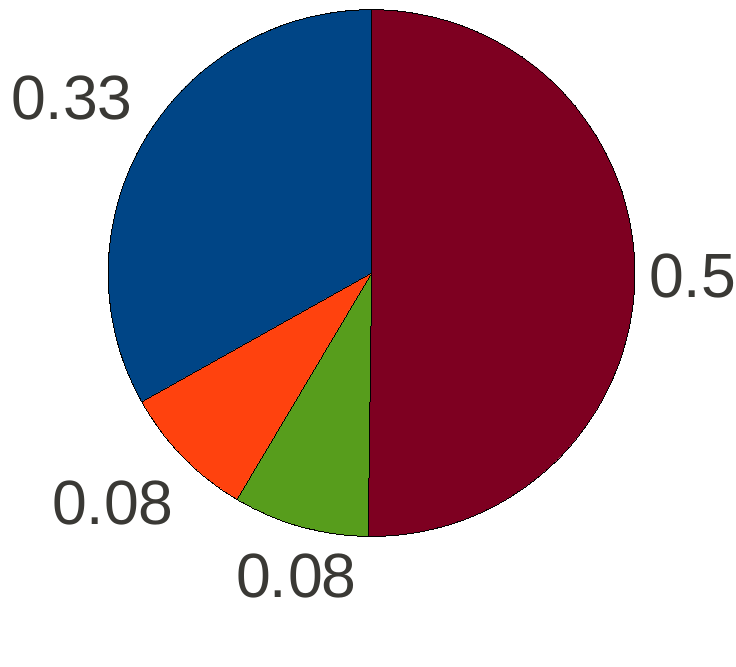} }~
	\subfloat[After iter. 2]{ \includegraphics[width=0.215\linewidth,trim=210 250 190 350,clip]{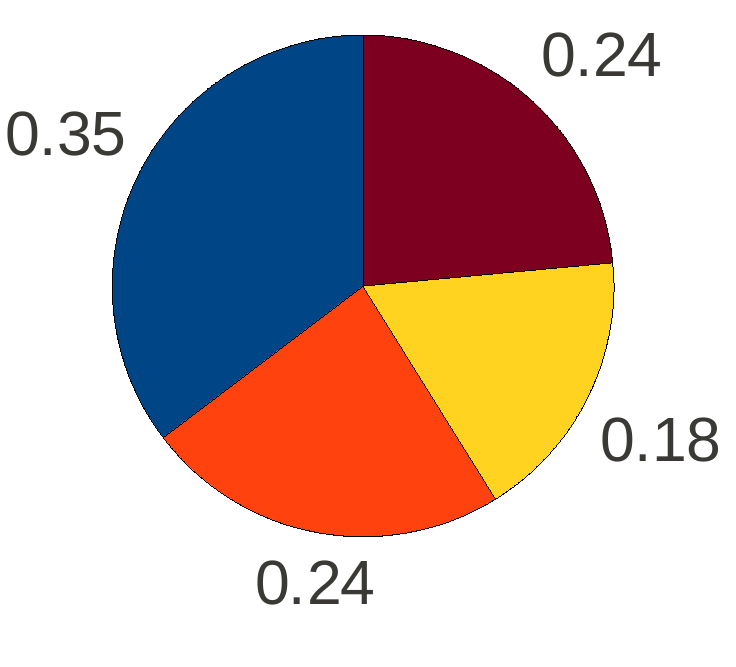} }~ 
	\subfloat[After iter. 3]{ \includegraphics[width=0.215\linewidth,trim=210 250 190 350,clip]{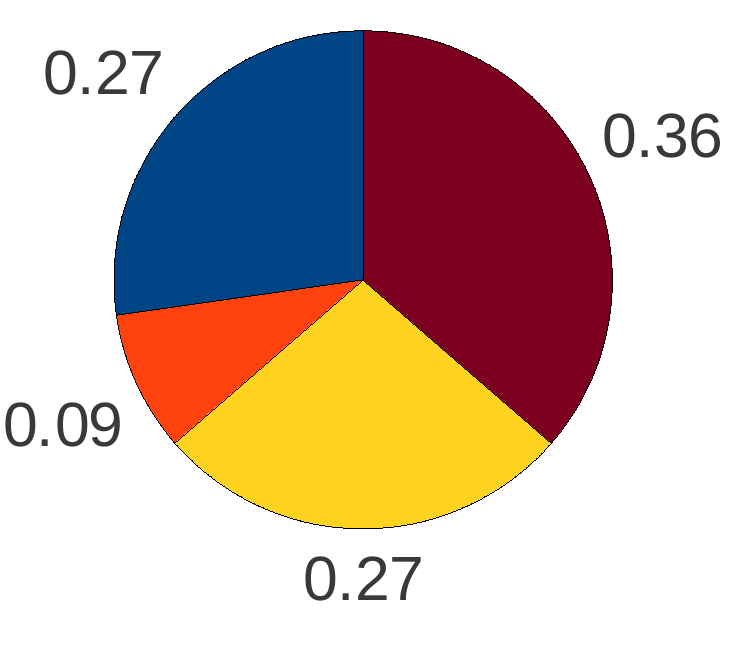} }
	\caption{Percentage of failures of each type for DAgger over the iterations of training in the high-density region.
	% Top (a-c): Low-Density Region, Bottom (d-g): High-Density Region.
	Blue: Large Trees, Orange: Thin Trees, Yellow: Leaves and Branches, Green: Other obstacles (poles, signs, etc.), Red: Too Narrow FOV. Clearly, a majority of the crashes happen due to a too narrow FOV and obstacles which are hard to perceive, such as branches and leaves.}
	\label{figTypeCrash}	
\end{figure}

As expected, we found that the narrow field-of-view was the largest contributor to failures of the reactive approach (Figure~\ref{dagger_failures}, right). The typical issue occurs when the learned controller avoids a tree, and as it turns a new tree comes into view. This may cause the controller to turn in a way such that it collides sideways into the tree it just avoided. This problem inevitably afflicts purely reactive controllers and could be solved by adding a higher level of reasoning \cite{bellingham2002receding}, or memory of recent visual features.

The type of failures are broken down by the type of obstacle the drone failed to avoid, or whether the obstacle was not in the FOV. Overall, 29.3\% of the failures were due to a too narrow FOV and 31.7\% on hard to perceive obstacles like branches and leaves.
%Figure \ref{figTypeCrash} compares the percentage of failures of each type over the iteration of training. The type of failures are broken down by the type of obstacle the drone failed to avoid (large tree, thin tree, or branches/leaves) when it was visible, and also failures where the obstacle was not clearly visible in the FOV of the drone.% In this figure, we can clearly see that a majority of the crashes were due to a too narrow FOV and hard obstacles to perceive like branches and leaves.

\begin{figure}%[!htbp]
	\centering
	\includegraphics[width=0.48\linewidth,trim=10 5 10 45,clip]{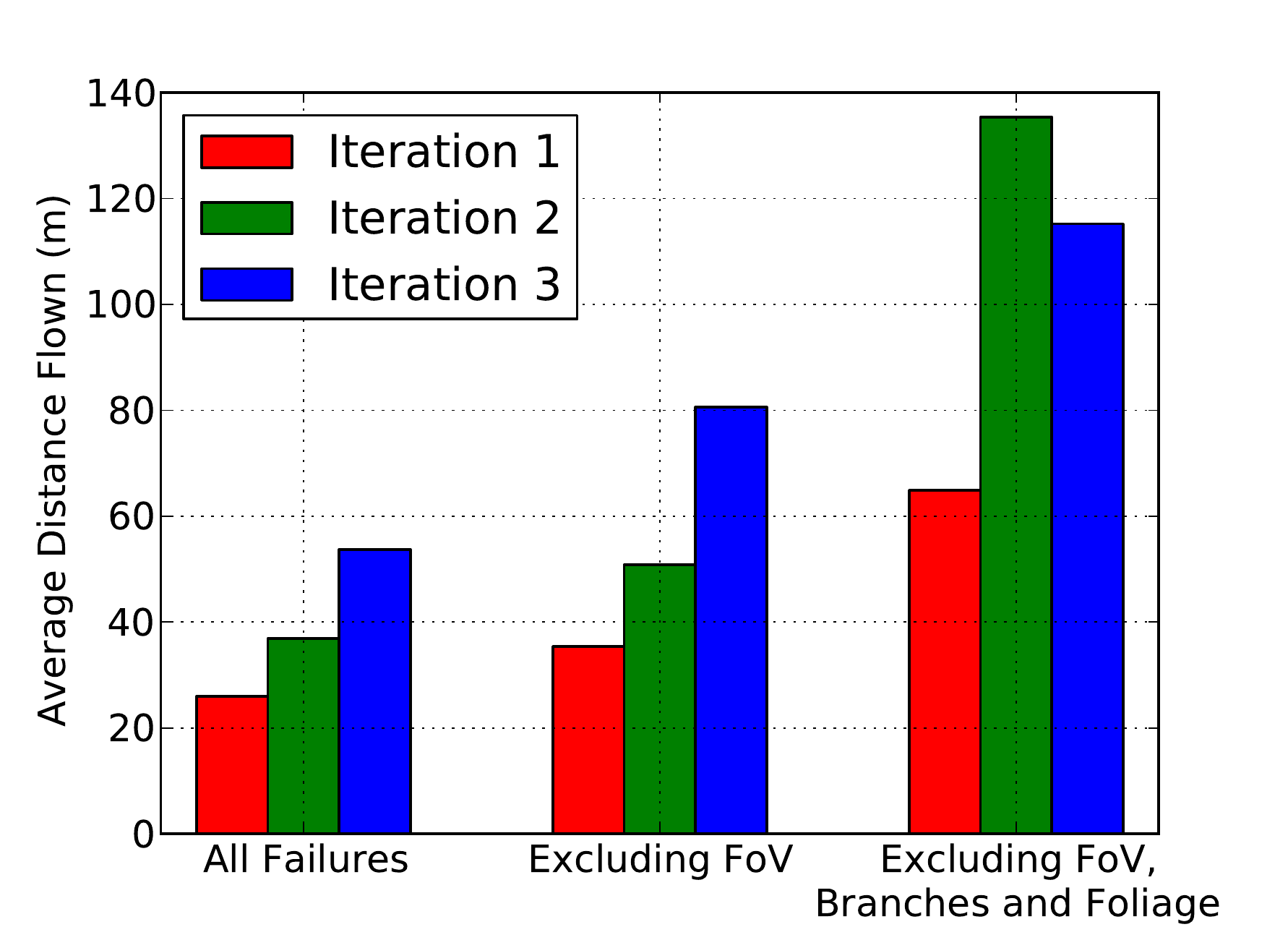}
        \includegraphics[width=0.48\linewidth,trim=10 5 10 45,clip]{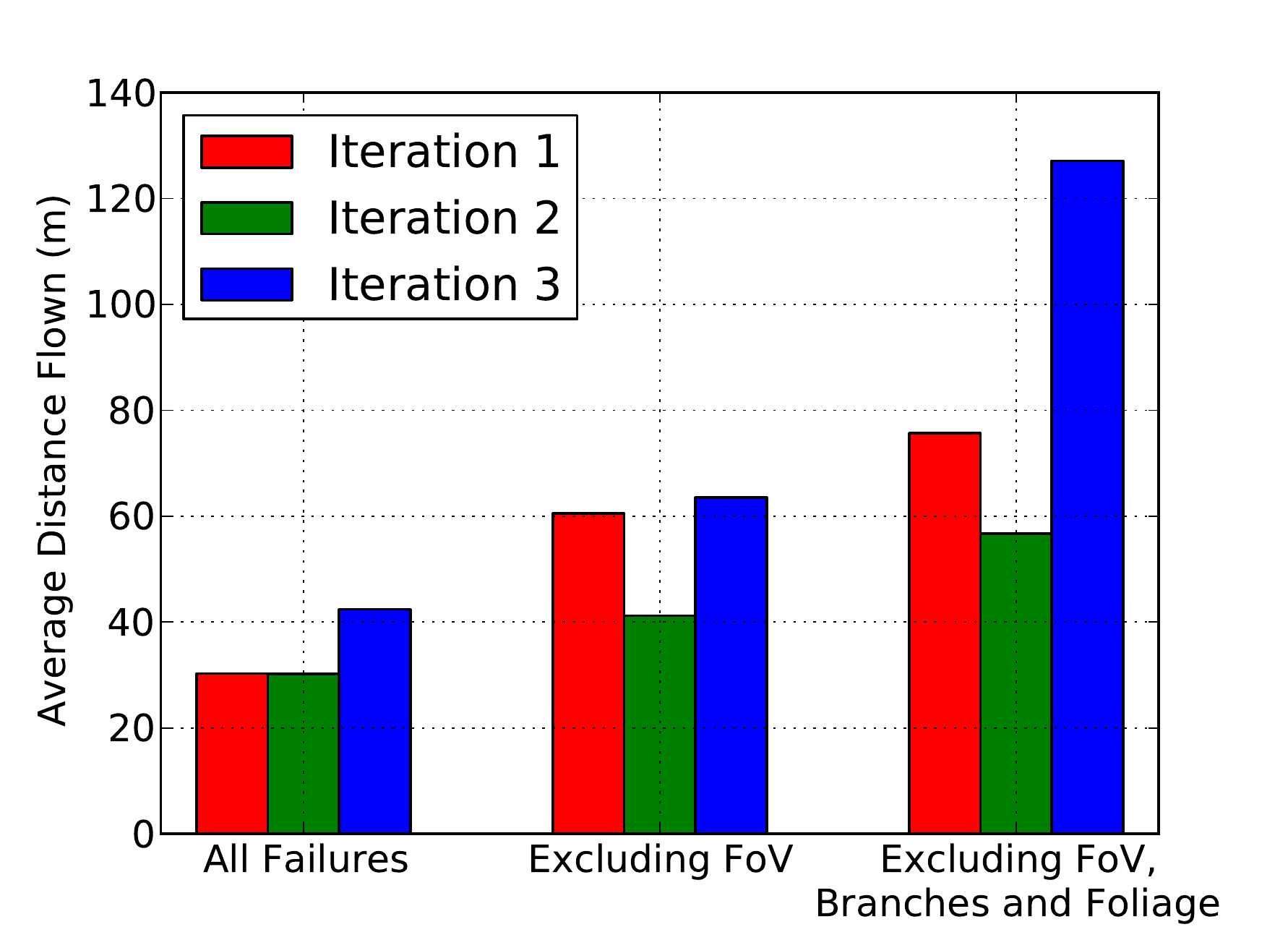}
	\caption{Average distance flown autonomously by the drone before a failure. Left: Low-Density Region, Right: High-Density Region. }
	\label{figDistance}
\end{figure}

Quantitatively, we compare the evolution of the average distance flown autonomously by the drone before a failure occured over the iterations of training. We compare these results when accounting for different types of failures in Figure \ref{figDistance} (left). When accounting for all failure types, the average distance flown per failure after all iterations of training was around $50$m. On the other hand, when only accounting for failures that are not due to the narrow FOV, or branches/leaves, the average distance flown increases to around $120$m. For comparison, the pilot successfully flew over 220m during the initial demonstrations, avoiding all trees in this sparse area. 

To achieve these results the drone has to avoid a significant number of trees. Over all the data, we counted the number of times the MAV avoided a tree\footnote{A tree is \textit{avoided} when the drone can see the tree pass from within its FOV to the edge of the image.}, and observed that it passed 1 tree every $7.5$m on average. We also checked whether the drone was actively avoiding trees by performing significant commands\footnote{A tree is \textit{actively avoided} when the controller issues a command larger than 25\% of the full range, passively in all other cases.}. 41\% of the trees were passed actively by our drone, compared to 54\% for the human pilot.

Finally, we tested whether the learned controller generalizes to new regions by testing it in a separate test area. The test area was slightly denser, around 1 tree per $10\times 10$m. The controller performed very well and was successfully able to avoid trees and perform at a similar level as in the training area. In particular, the drone was able to fly autonomously without crashing over a $100$m distance, reaching the limit of our communication range for the tests.

%For the outdoor experiments the drone was operated autonomously over ***km for a net autonomous flight time duration of *** minutes. Our training dataset comprised two categories of test regions - a sparser park and a dense wooded region with a tree density of 0.8 trees per 10 square meters. 

%For the purpose of classification of data regarding avoiding trees, the human labellers adhered to the following criteria 
%\begin{itemize}
%\item 
%\item A \textit{useful input} is a velocity command (or a series of smaller continuous commands) output by DAgger above a threshold depending on current forward velocity which causes the drone to significantly move to avoid a tree. No condition is imposed on the input being successful in avoiding a tree.
%\item A \textit{non useful input} is correspondingly any command velocity greater than the above mentioned threshold that does not assist in avoiding a tree. Examples would be the drone turning while in open terrain and heading into a tree after an initial command of heading away
%\item Trees are passed through the middle if the separation between the two trees is within a three meters and the drone can see both trees in its field of view moving towards the corresponding edges. These trees are not counted in the trees avoided by moving left or right.

%***Mention two sets of tests, a sparser training region and a denser region, %and develop the argument as a continuous progress from a slower speed and %lower resolution to a greater forward speed and higher camera resolution***

%\end{itemize}

\subsubsection{High-density test region}

The second set of experiments was conducted in a thickly wooded region in a
local forest. The tree density was significantly higher, around 1 tree per
$3\times 3$m, and the area included a much more diverse range of trees,
ranging from very small and thin to full-grown trees. In this area we flew at
a faster fixed velocity of around $1.5$m/s, and again learned the heading
(left-right) controller to avoid trees using DAgger over 3 iterations of
training. This represented a total of $1.2$km of flight training data. The
final controller was also tested over additional $400$m of flight in this
area. For this experiment however, we used the new ARDrone 2.0 quad-rotor
helicopter, which has an improved HD camera that can stream $640\times 360$
pixel images at $30$Hz. The increased resolution probably helped in detecting the thinner trees. 

Qualitatively, in this experiment we observed that the performance of the learned behavior slightly decreased in the second iteration, but then improved significantly after the third iteration. For example, we observed more failures to avoid both thin and large trees in the second iteration compared to the other iterations. This is shown in Figure \ref{figTypeCrash}, which compares the percentage of the different failures for the human pilot and after each iteration of DAgger in this area. We can also observe that the percentage of failures attributed to large or thin trees is smallest after the third iteration, and that again a large fraction of the failures occur when obstacles are not visible in the FOV of the MAV. Additionally, we can observe that the percentages of failures due to branches or leaves diminishes slightly over the iterations, which could be attributed to the better camera that can better perceive these obstacles.  A visualization of a typical sequence is given in Figure~\ref{fig:flight_visualization}. Further qualitative results can be found in the supplementary material.

\begin{figure}%[!htbp]
    \centering
	\begin{tabular}{cc}
	             \includegraphics[width=0.45\linewidth]{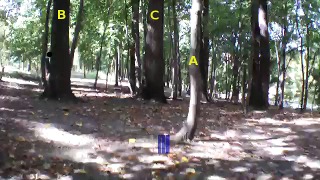}  & \includegraphics[width=0.45\linewidth]{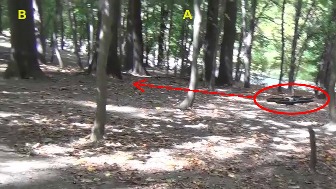} \\
	             \includegraphics[width=0.45\linewidth]{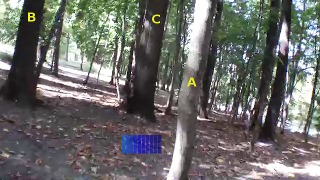}  & \includegraphics[width=0.45\linewidth]{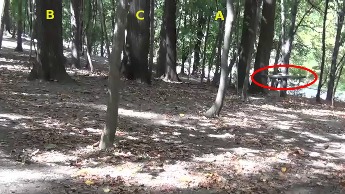}\\
	             \includegraphics[width=0.45\linewidth]{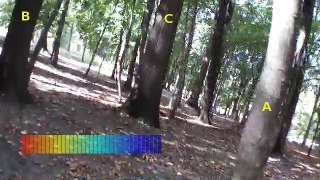}  & \includegraphics[width=0.45\linewidth]{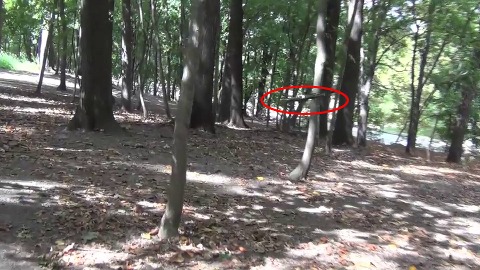}\\
	             \includegraphics[width=0.45\linewidth]{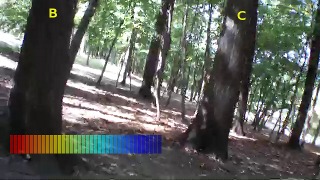}  & \includegraphics[width=0.45\linewidth]{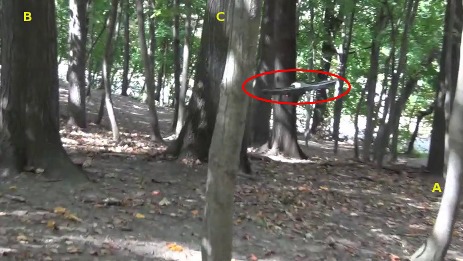}\\
	             \includegraphics[width=0.45\linewidth]{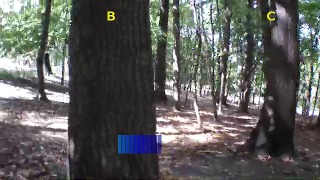}  & \includegraphics[width=0.45\linewidth]{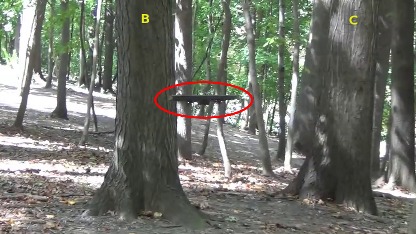}\\
	             \includegraphics[width=0.45\linewidth]{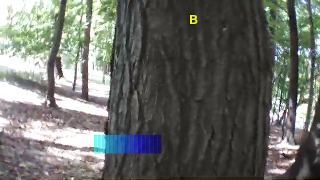}  & \includegraphics[width=0.45\linewidth]{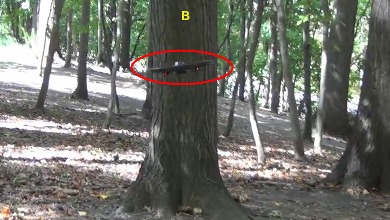}\\
	             \includegraphics[width=0.45\linewidth]{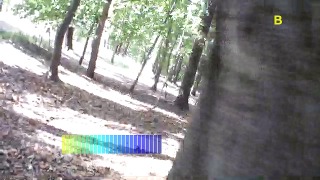}  & \includegraphics[width=0.45\linewidth]{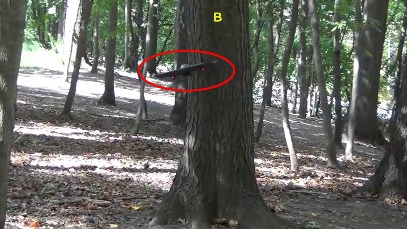}\\
                 \includegraphics[width=0.45\linewidth]{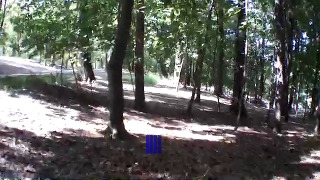} & \includegraphics[width=0.45\linewidth]{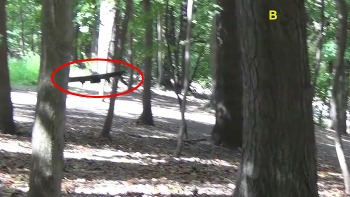} \\
	             MAV's on-board view & Observer's view
    \end{tabular}
	\caption{Example flight in a dense forest area. The image sequence is chronologically ordered from top ($t=0s$) to bottom ($t=6.6s$) and split into the MAV's on-board view on the left and an observer's view to the right. Note the direction label of the MAV in the first frame, and the color-coded commands issued by DAgger. Apart from a qualitative visualization of our algorithm, two major things can be observed. First, after avoiding tree \emph{A} in frame 3 the vehicle still rolls strongly to the left in frame 4. This is due to the small but ubiquitous latency and should be addressed in future work to fly the MAV in even denser areas. Second, in frames 5-7 tree \emph{B} is avoided on the left, rather than on the more intuitive right. DAgger prefers this decision based on the drift feature, which indicates that the vehicle still moves left and thus a swerve to the right would be more difficult.}
	\label{fig:flight_visualization}
\end{figure}

Quantitatively, we compare the evolution of the average distance flown autonomously by the MAV before a failure occurred over the iterations of training. Again, we compare these results when accounting for different types of failures in Figure \ref{figDistance} (right). When accounting for all failure types, the average distance flown per failure after all iterations of training was around $40$m. Surprisingly, despite the large increase in tree density and faster forward velocity, this is only slightly worse than our previous results in the sparse region. Furthermore, when only accounting for failures that are not due to the narrow FOV or branches and leaves, the average distance flown increases to $120$m per failure, which is on par with our results in the sparser area. For comparison, when only accounting for failures due to tree trunks, the pilot flew around 500m during the initial demonstrations and only failed to avoid one thin tree. However, the pilot also failed to avoid thin branches and foliage more often (Figure \ref{figTypeCrash}). When accounting for all types of obstacles, the pilots average distance until failure was around $80$m.

The increase in tree density required our MAV to avoid a significant larger number of trees to achieve these results. Over all the data, we observed that it was avoiding on average 1 tree every 5m. In this dense region, both the pilot and the drone  had to use larger controls to avoid all the trees, leading to an increase in the proportions of trees that were passed actively. 62\% of the trees were passed actively by the drone, compared to a similar ratio of 66\% for the pilot.

%\subsubsection{Discussion of the Outdoor Results}

The major issue of the too narrow FOV presented in this section may be addressed by two approaches in the future. First, imitation learning methods that integrate a small amount of memory may allow to overcome the simplest failure cases without resorting to a complete and expensive mapping of the environment. Second, the biologically-inspired solution is to simply ensure a wider FOV for the camera system. For example, pigeons rely mostly on monocular vision and have a FOV more than 3 times larger, while owls have binocular vision with around 1.5 times the FOV of our drone.% Hence, a case can be made for requiring wider FOV.

%%%%%%%%%%%%%%%%%%%%%%%%%%%%%%%%%%%%%%%%%%%%%%%%%%%%%%%%%%%%%%%%%%%
\section{Conclusion}
\label{sec.conclusion}

We have presented a novel approach for high-speed, autonomous MAV flight through dense forest environments. Our system learns to predict how a human expert would control the aircraft in a similar situation, and thus successfully avoids collisions with trees and foliage using passive, low-cost and low-weight visual sensors only. We have applied a novel imitation learning strategy which takes into account that the MAV is likely to end up in situations where the human pilot does not, and thus needs to learn how to react in such cases. During a significant amount of outdoor experiments with flights over a distance of  $3.4$km, our approach has avoided more than $680$ trees in environments of varying density.

% Mention purpose of reactive control again!
%This is an important feature, as the purpose of any reactive controller is to provide a reliable, low-level layer for autonomous control, which works on minimal visual input and can handle situations where 3D mapping~\cite{achtelik2011onboard, wendel2012dense} or high-level trajectory planning~\cite{bellingham2002receding} fails.

Our work provides an important low-level layer for autonomous control of MAVs, which works on minimal visual input and can handle situations where 3D mapping or high-level trajectory planning fails. In future work, we want to focus on adding such higher-order layers, including receding horizon planning and semantic knowledge transfer, as well as implementing the means to handle latency and small field of view effects. This will allow us to perform even longer flights, in even denser forests and other cluttered environments.
% Our novel method is evaluated in a constrained indoor setting using motion capture, as well as in several forest environments. In total, we successfully avoided more than $900$ trees during flights over a distance of more than $3$km, which proves our approach to be highly effective.

% Future work
%We plan to extend this work on feature computaiton for reactive control in several ways. Although the computation time is acceptable for moderate to slow speeds, we need to move to faster implementation, such as FPGA for faster flights. The features are currently low-level features which do not incorporate any semantic knowledge about the environment. We need to include the semantic labeling output into the reactive control loop. 

%%%%%%%%%%%%%%%%%%%%%%%%%%%%%%%%%%%%%%%%%%%%%%%%%%%%%%%%%%%%%%%%%%%

\section*{Acknowledgment}

This work has been supported by ONR through BIRD MURI.
A. Wendel acknowledges the support of the Austrian Marshallplan Foundation during his research visit at CMU.



%\bibliographystyle{IEEEtran}

\bibliographystyle{IEEEtran}
\bibliography{IEEEabrv,references}

\end{document}